\documentclass{amia}
\usepackage{lipsum} 
\usepackage{multirow}
\usepackage{amssymb}
\usepackage{amsmath}
\usepackage{color}
\begin{document}

\title{Multi-Task Learning for Post-transplant Cause of Death Analysis: A Case Study on Liver Transplant}

\author{Sirui Ding$^1$, Qiaoyu Tan$^1$, Chia-yuan Chang$^1$, Na Zou, PhD$^1$, Kai Zhang, PhD$^3$, Nathan R. Hoot, MD, PhD$^4$ Xiaoqian Jiang, PhD$^3$, Xia Hu, PhD$^2$ }

\institutes{
    $^1$ Texas A\&M University, College station, TX, USA; $^2$Rice University, Houston, TX, USA; $^3$University of Texas Health Science Center, Houston, TX, USA; $^4$McGovern Medical School, University of Texas Health Science Center, Houston, TX, USA. 
}

\maketitle

\section*{Abstract}

\textit{Organ transplant is the essential treatment method for some end-stage diseases, such as liver failure. Analyzing
the post-transplant cause of death (CoD) after organ transplant provides a powerful tool for clinical decision making,
including personalized treatment and organ allocation. However, traditional methods like Model for End-stage Liver Disease (MELD) score and conventional
machine learning (ML) methods are limited in CoD analysis due to two major data and model-related challenges. To address this, we propose a novel framework called CoD-MTL leveraging multi-task learning to model the semantic relationships between various CoD prediction tasks jointly.
Specifically, we develop a novel
tree distillation strategy for multi-task learning, which combines the strength of both the tree model and multi-task learning. Experimental results are presented to show the precise and reliable CoD predictions of our framework.
A case study is conducted to demonstrate the clinical importance of our method in the liver transplant.}

\section{Introduction}
Organ transplant is a crucial therapeutic option for individuals with end-stage diseases, e.g., kidney failure~\cite{abecassis2008kidney}, liver failure~\cite{kumar2021liver}, liver cancer~\cite{ravaioli2014liver}, etc. However, due to the complex surgical procedures and high risk of graft failure~\cite{tovikkai2015time}, how to allocate organs properly remains an important yet challenging problem. To increase allocation precision and effectiveness, doctors often need to consider a series of post-transplant factors, especially the cause of death (CoD) analysis~\cite{watt2010evolution}, such as rejection, infection, cancer, and recurrent disease~\cite{moreno2006post}. Accurately predicting and analyzing these CoDs before the transplant can aid doctors in making better clinical decisions regarding organ allocation~\cite{volk2015decision} and precise treatment after the surgery~\cite{bhat2014care}. In this work, we focus on liver transplant as a case study. Currently, the MELD score~\cite{wiesner2003meld} is widely used as the standard medical indicator to aid doctors in making better clinical decisions. Nevertheless, MELD cannot provide a granular analysis of the aforementioned CoDs factors, since it was originally designed for the 3-month mortality prediction of liver-related diseases. While some statistical methods have been proposed, they are either intended for a limited number of predictors~\cite{gong2020predictors} or make strong assumptions about the input features and outcomes, such as linear relations and feature independence~\cite{nitski2021long}. These limitations hinder the accurate prediction of post-transplant CoDs, necessitating the development of more advanced computational methods to support precise clinical decision-making in liver transplant.   


Machine learning (ML) has recently received remarkable success in predicting transplant-related medical outcomes~\cite{gotlieb2022promise}. For example, Lau et al. employed neural networks and random forest to predict post-transplant graft failure~\cite{lau2017machine}. Ding et al. developed a prediction framework based on knowledge distillation for the graft status prediction with consideration of fairness issues~\cite{ding2023fairly}. Despite their success, the complex nature of liver transplant makes it infeasible to apply previous ML methods directly for post-transplant CoD prediction. We identify two significant challenges from the data and model-related aspects as follows.

First, from a data perspective, a patient usually has multiple CoDs which makes the analysis a multi-label learning task. In addition, recorded CoDs (positive samples) are scarce compared to negative samples, i.e., successful transplantation or unrecorded data. As a result, there is an imbalance problem in the data, making it difficult for machine learning models to accurately predict the positive class~\cite{kaur2019systematic}. This is because we do not have enough data to learn ML models for different CoD tasks independently. 
Therefore, it is infeasible to directly apply traditional multi-class learning methods~\cite{li2018multi} and existing ML methods for organ transplant~\cite{gotlieb2022promise} in the post-transplant CoDs analysis.

Second, from a modeling perspective, tree-based models like GBDT~\cite{friedman2001gbdt,sayed2021predicting} tend to perform better than neural network (NN) based approaches~\cite{shwartz2022tabular} in the healthcare field, since the majority of organ transplantation records are EHR$/$tabular data~\cite{massie2014big}. 
We also verified this in our preliminary experiments, as shown in Table~\ref{mainexp}. Despite the relative advantages of tree-based models, they are still limited in tackling our CoD tasks, since they cannot capture the complementary correlations among different CoD tasks (a.k.a. multiple labels)~\cite{zhang2020gbdt}. Thus, there is an urgent need to devise more advanced tree-aware models that can simultaneously handle multiple prediction targets. 


To tackle the above challenges, we propose a tree-distillation multitask learning framework, called \textbf{CoD-MTL}, for post-transplant CoD analysis. In this paper, we focus on the prediction of rejection and infection since they are the most common post-transplant CoDs~\cite{watt2010evolution}. Specifically, for challenge (1), instead of modeling the rejection and infection independently, we develop a multitask learning model~\cite{zhang2021survey} with a shared network layer under the CoD-MTL framework to capture their semantic correlations, since they are intrinsically associated with each other in the organ transplantation field. The shared neural networks will take advantage of the various related tasks to alleviate the unbalanced data problem in CoD analysis. For challenge (2), we design a novel tree distillation strategy in CoD-MTL to effectively transfer the advances of tree-based models into neural networks for different CoD tasks. As a result, a principled approach is obtained to integrate the capacity of multitask learning in capturing complementary information across various tasks and the power of tree-based models in modeling tabular data in an end-to-end fashion.
We validate the effectiveness of our framework on the real-world liver transplant dataset. Experiment results show the CoD-MTL can accurately predict the post-transplant CoDs. The case study demonstrates the clinical importance of CoD-MTL to help doctors in organ transplant clinical decisions.

\begin{figure*}[t]
  \centering
    \includegraphics[width=1.0\textwidth]{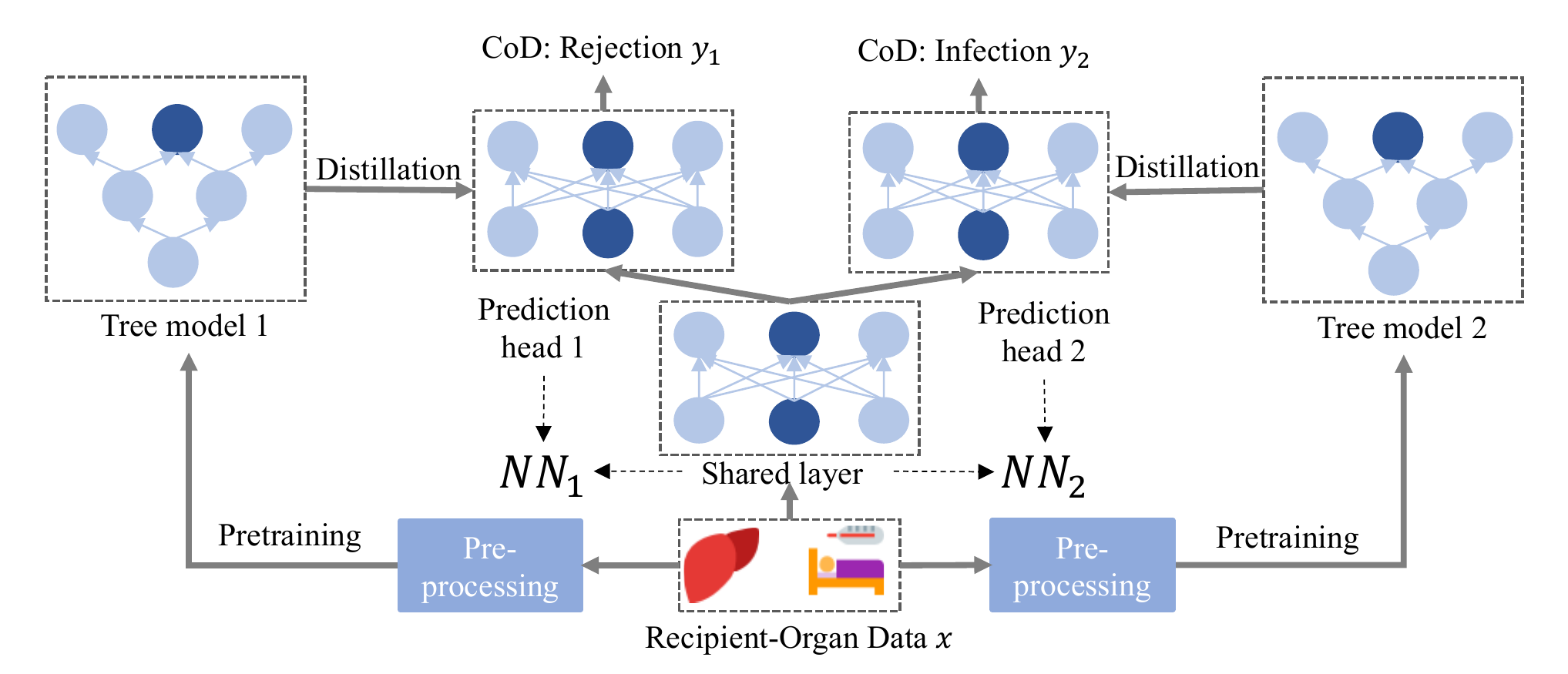}
 \caption{An overview of the CoD-MTL workflow for multiple CoDs prediction.}
 \label{fig:workflow}
\end{figure*}

\section{Data and Problem Description}

\textbf{Data preparation.} In this work, we use a patient cohort obtained from the patients registered on the liver transplant waiting list of the Organ Procurement and Transplantation Network (OPTN)~\cite{kwong2020optn}, consisting of a total of $8,922$ patients who underwent liver transplantation. Out of these patients, $4,160$ died due to rejection (including both acute and chronic rejection), and $3627$ died due to infection after the transplant. In addition, we also randomly selected 2000 patients as negative samples who had no documented death after transplantation.  
In this study, we consider $102$ features from both the donor organs and recipients, excluding sensitive attributes such as gender and race. The donor/organ features are divided into three categories: the donor's basic information, the donor's history of diseases, and information on the donor's death. Similarly, the patient/recipient features are categorized as the patient's basic information, history of diseases, and transplant-related laboratory tests.

\textbf{Problem formulation.}
We are given a dataset $\{p_i, o_i, Y_i\}_{i=1}^N$ consisting of $N$ patient-organ pairs. Each patient $p_i$ (or organ $o_i$) is associated with a $L_p$ (or $L_o$) dimensional feature vector $\mathbf{x}_{i}^p\in\mathbb{R}^{L_p}$ (or $\mathbf{x}_i^o\in\mathbb{R}^{L_o}$).
For each patient-organ pair $(p_i, o_i)$, there are $M$ possible causes of death (CoDs), denoted as $Y_i = \{y_j\in\{0,1\}\}_{j=1}^M$, where $y_j=1$ if the $j$-th CoD causes the death of the patient and $y_j=0$ otherwise. 
The goal is to train a machine learning model that can predict the probability of each CoD for a given patient-organ pair $(p_i, o_i)$ based on their input features $\mathbf{x}_i = \langle \mathbf{x}^p_i, \mathbf{x}^o_i \rangle$. The model should learn to predict multiple CoDs simultaneously, and the learning objective is to minimize the cross-entropy loss between the predicted probabilities and the ground-truth labels.

\section{Methodology}
This section will introduce the proposed post-transplant CoD prediction framework (CoD-MTL) in detail. Firstly, we will describe the pre-processing procedure for input data (Section~\ref{datapreprocessing}). Then we introduce the multi-task learning framework for post-transplant CoDs prediction (Section~\ref{overviewofworkflow}). Finally, the proposed tree-distillation strategy for multi-task learning will be elaborated (Section~\ref{treediststra}). 

\subsection{Data pre-processing}
\label{datapreprocessing}
To effectively learn from original liver transplant EHR data, we use an encoder to transform categorical features into numerical values following the standard ways of processing raw data. These numerical features are then concatenated with the original numerical features. To address any missing values, we impute all features with zero. This processed data is used as input for both the tree and multi-task learning models in the CoD-MTL framework. Additionally, to ensure a robust evaluation, the data samples are shuffled during the K-fold cross-validation stage.

\subsection{Multi-task learning for multiple CoDs prediction}
\label{overviewofworkflow}
Immunosuppressive drugs that patients take to prevent rejection after liver transplant surgery can weaken their immune system and increase their susceptibility to infections~\cite{hernandez2015infectious}. To investigate the clinical relationship between rejection and infection prediction tasks, we adopt a multi-task learning approach that uses a shared deep learning module and customized prediction heads for different CoDs. The CoD-MTL framework is designed based on the multi-task learning paradigm~\cite{caruana1998multitask}, as illustrated in Figure~\ref{fig:workflow}. To predict the $j$-th CoD, we formulate the output as follows:

\begin{equation}
    y_j = \text{Head}_j(\text{SharedLayer}(x_i)),
    \label{overallheadshared}
\end{equation}

where $\text{Head}_j(\cdot)$ refers to the prediction head part for the $j$-th task, and $\text{SharedLayer}(\cdot)$ denotes the shared layer of multiple tasks in the multi-task learning pipeline. We will provide further details about the $\text{Head}$ and $\text{SharedLayer}$ modules in the following subsection.

\subsection{Tree-distillation boosted multi-task learning}
\label{treediststra}

In this subsection, we will elaborate on the proposed tree-distillation strategy in the multi-task learning framework. Firstly, we will present the process of integrating the tree model into neural networks using knowledge distillation. Next, we will introduce a new approach to integrate the tree models into the multi-task learning framework. Then we will describe the learning process of the whole CoD-MTL framework.


\textbf{Tree distillation in the neural network.} Tree-based models like GBDT have shown great success across various healthcare scenarios and tabular data~\cite{chang2019machine,seto2022gradient}. Recently, DeepGBM~\cite{ke2019deepgbm} has been developed to combine the merits of GBDT and deep neural networks by distilling the knowledge of GBDT to deep neural networks. Despite its effectiveness, DeepGBM is designed for a single learning task and cannot model the correlations between multiple learning tasks, as shown in CoD analysis. Inspired by this, we propose to upgrade DeepGBM for multi-task learning, i.e., distilling multiple task-specific GBDT models into a multi-task deep neural network. Assume $V^{t,i}$ denotes the sparse leaf index that corresponds to the $i$-th patient of the training data in the $t$-th tree of $T$, we first transform the leaf outputs of one GBDT model $T$ into a dense embedding as below:

\begin{equation}
    \mathbf{E}^i = \text{Emb}(||_{t\in T}(V^{t,i}); \theta),
    \label{emb}
\end{equation}

where $\mathbf{E}^i$ represents the dense embedding table obtained from the embedding model $\text{Emb}(\cdot)$ with trainable parameter $\theta$, where $\text{Emb}(\cdot)$ is a fully connected neural network. The notation $||_{t\in T}(V^{t,i})$ indicates the concatenated sparse representation across multiple trees in GBDT. To learn the embedding model, we optimize the objective function as:

\begin{equation}
    min \frac{1}{N} \sum_{i=1}^{N} \mathcal{L'}(\mathbf{W}\times \text{Emb}(||_{t\in T}(V^{t,i}); \theta) + \textbf{b}, q^i),
    \label{emb}
\end{equation}

where $\textbf{W}$ and $\textbf{b}$ are the parameters that map the dense embedding into the final prediction, and $q^i$ is the corresponding leaf prediction of the $i$-th sample. The loss function $\mathcal{L'}$ can be chosen as the cross-entropy loss function, which is commonly used in classification tasks.

After the embedding of sparse representations from tree models' leaves, we can use this dense embedding $E^i$ as the distillation target to further distill the tree structures into a neural network. The distilled neural network can approximate the tree model by optimizing the following objective:

\begin{equation}
    \mathcal{L}_{distill} = \frac{1}{N} \sum_{i=1}^{N} \mathcal{L}(\text{NN}(\mathbf{x}^p_i [\mathbb{I}^T]); \theta_{NN}), \mathbf{E}^i),
    \label{emb}
\end{equation}

where $\text{NN}(\cdot)$ represents the distilled neural network with trainable parameters $\theta_{NN}$, and $\mathbf{x}_i^p$ is the input feature for the $i$-th patient. $\mathbb{I}$ denotes the indices of the features selected from the tree model.


\begin{table*}[t]
\centering
\begin{tabular}{|c|c|cc|cc|}
\hline
\multirow{2}{*}{\textbf{Model type}}          & \multirow{2}{*}{\textbf{Model}} & \multicolumn{2}{c|}{\textbf{CoD: Rejection}}         & \multicolumn{2}{c|}{\textbf{CoD: Infection}}         \\ \cline{3-6} 
                                              &                                 & \multicolumn{1}{c|}{\textbf{AUROC}} & \textbf{AUPRC} & \multicolumn{1}{c|}{\textbf{AUROC}} & \textbf{AUPRC} \\ \hline
\multirow{3}{*}{Traditional ML (single task)} & Logistic Regression             & \multicolumn{1}{c|}{0.551+0.008}    & 0.482+0.005    & \multicolumn{1}{c|}{0.569+0.013}    & 0.471+0.013    \\ \cline{2-6} 
                                              & GBDT                            & \multicolumn{1}{c|}{0.588+0.008}    & 0.497+0.010    & \multicolumn{1}{c|}{0.611+0.011}    & 0.499+0.014    \\ \cline{2-6} 
                                              & Random Forest                   & \multicolumn{1}{c|}{0.583$\pm$0.016}    & 0.504$\pm$0.009    & \multicolumn{1}{c|}{0.608$\pm$0.009}    & 0.506$\pm$0.020    \\ \hline
Neural Networks (single task)                 & MLP                             & \multicolumn{1}{c|}{0.571+0.012}    & 0.493+0.008    & \multicolumn{1}{c|}{0.592+0.003}    & 0.483+0.011    \\ \hline
Multitask learning model                      & Multitask Learning              & \multicolumn{1}{c|}{0.595+0.021}    & 0.517+0.015    & \multicolumn{1}{c|}{0.614+0.019}    & 0.515+0.028    \\ \hline
The proposed method                           & CoD-MTL                         & \multicolumn{1}{c|}{0.640+0.012}    & 0.557+0.012    & \multicolumn{1}{c|}{0.646+0.007}    & 0.553+0.018    \\ \hline
\end{tabular}
\caption{Performance comparison on Two CoD Prediction Tasks}
\label{mainexp}
\end{table*}

\textbf{Integration of tree model in multi-task learning.} When it comes to predicting multiple post-transplant CoDs, we propose a multi-task tree-distillation paradigm to achieve this. First, we train a GBDT model for each CoD prediction task. For the $j$-th CoD, we have GBDT model $T_j$ and the distilled network $\text{NN}_j(\cdot)$ with trainable parameters $\theta_{NN_j}$. We then develop the distilled neural network for multiple CoD tasks, as shown in Figure~\ref{fig:workflow}. Specifically, the distilled model $NN_j$ for each CoD task includes a shared layer for representation learning and a task-specific prediction head for each CoD task, as shown in Formula~\ref{overallheadshared}.
The prediction head for the $j$-th CoD is a simple neural network as follows.

\begin{equation}
    y_j(x^i) = \mathbf{W}_j \times NN_j(\mathbf{x}_i [\mathbb{I}^{T_j}]); \theta_{NN_j}) + \mathbf{b}_j,
    \label{emb}
\end{equation}

where $\mathbf{W}_j, \mathbf{b}_j$ are associated parameters to transfer the dense embedding to final predictions for the $j$-th task.

\textbf{Learning process of CoD-MTL.} To train our model, we optimize the parameters of CoD-MTL according to the following multi-task loss function. 

\begin{equation}
\begin{aligned}
    \mathcal{L}_{j} = \frac{1}{N} \sum_{i=1}^{N} \mathcal{L}(\text{NN}_j(\mathbf{x}_i [\mathbb{I}^{T_j}]); \theta_{NN_j}), \mathbf{E}_j^i),
    \\
    \mathcal{L}_{multi} = \sum_{j=1}^{M} \alpha_j(\beta_j \mathcal{L}'(y_j, y'_j)+ \gamma_j \mathcal{L}_j).
\end{aligned}
    \label{emb}
\end{equation}

$\mathcal{L}_j$ is the knowledge distillation loss function for the $j$-th CoD task, i.e, $NN_j$. $\mathcal{L}_{multi}$ is the overall multi-task loss function for $M$ CoD tasks, where 
$\alpha_j, \beta_j$, and $\gamma_j$ are trade-off parameters to control the importance of different terms. 

\section{Experiment}

\begin{figure}[]
  \centering
    \includegraphics[scale=0.6]{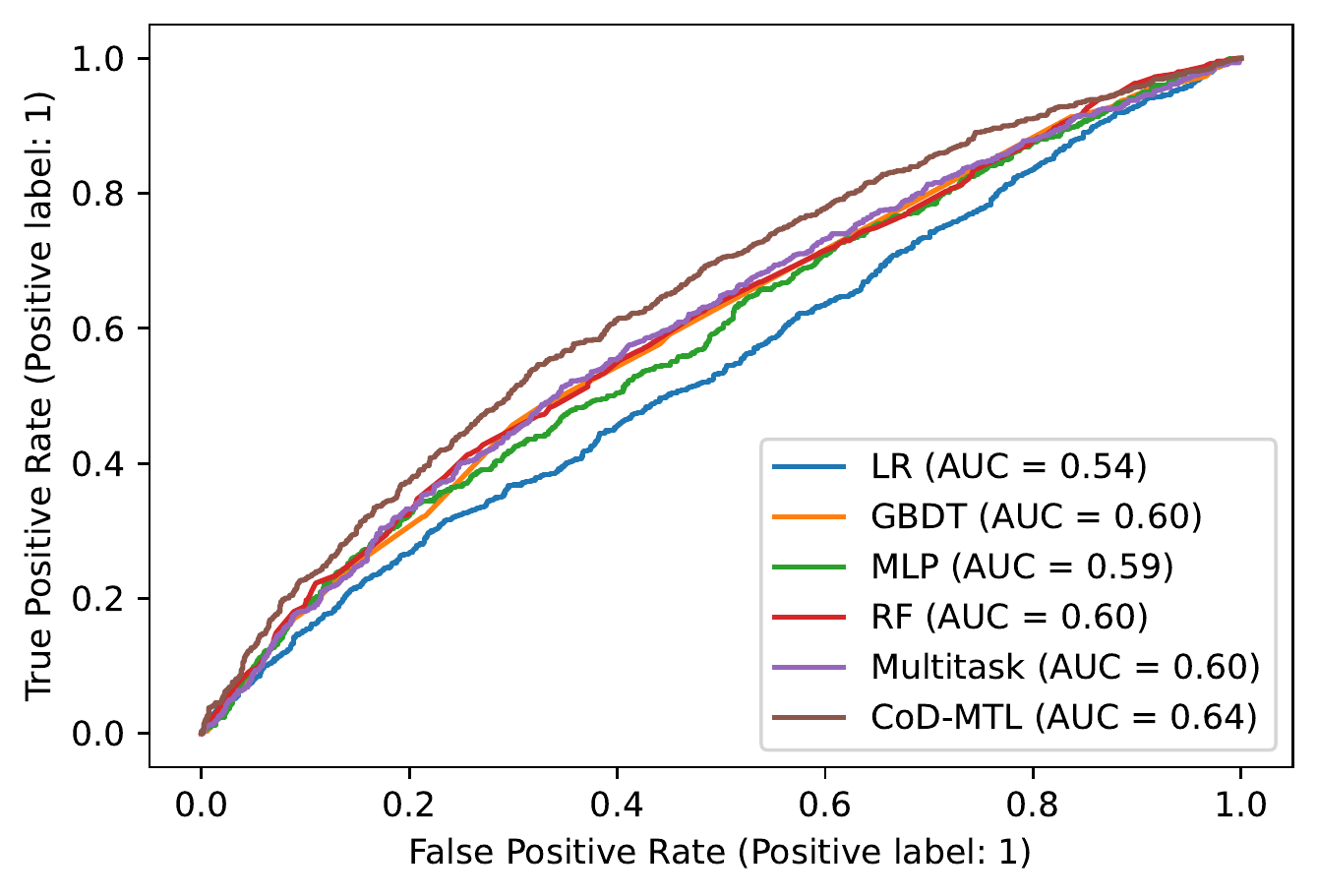}
    \includegraphics[scale=0.6]{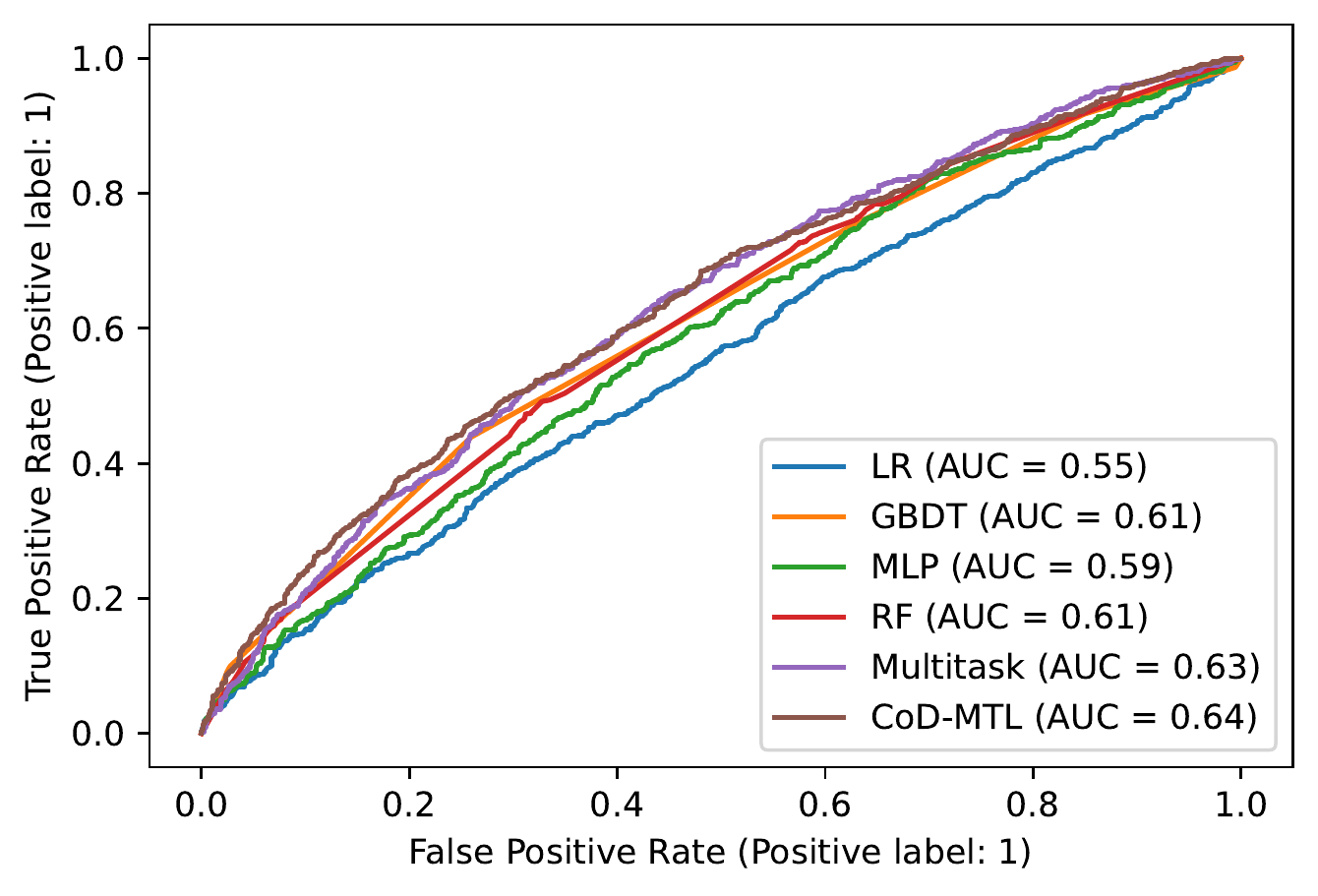}
 \caption{ROC curves for rejection and infection CoDs (From left to right).}
 \label{fig:roccurve}
\end{figure}

\begin{figure}[]
  \centering
    \includegraphics[scale=0.6]{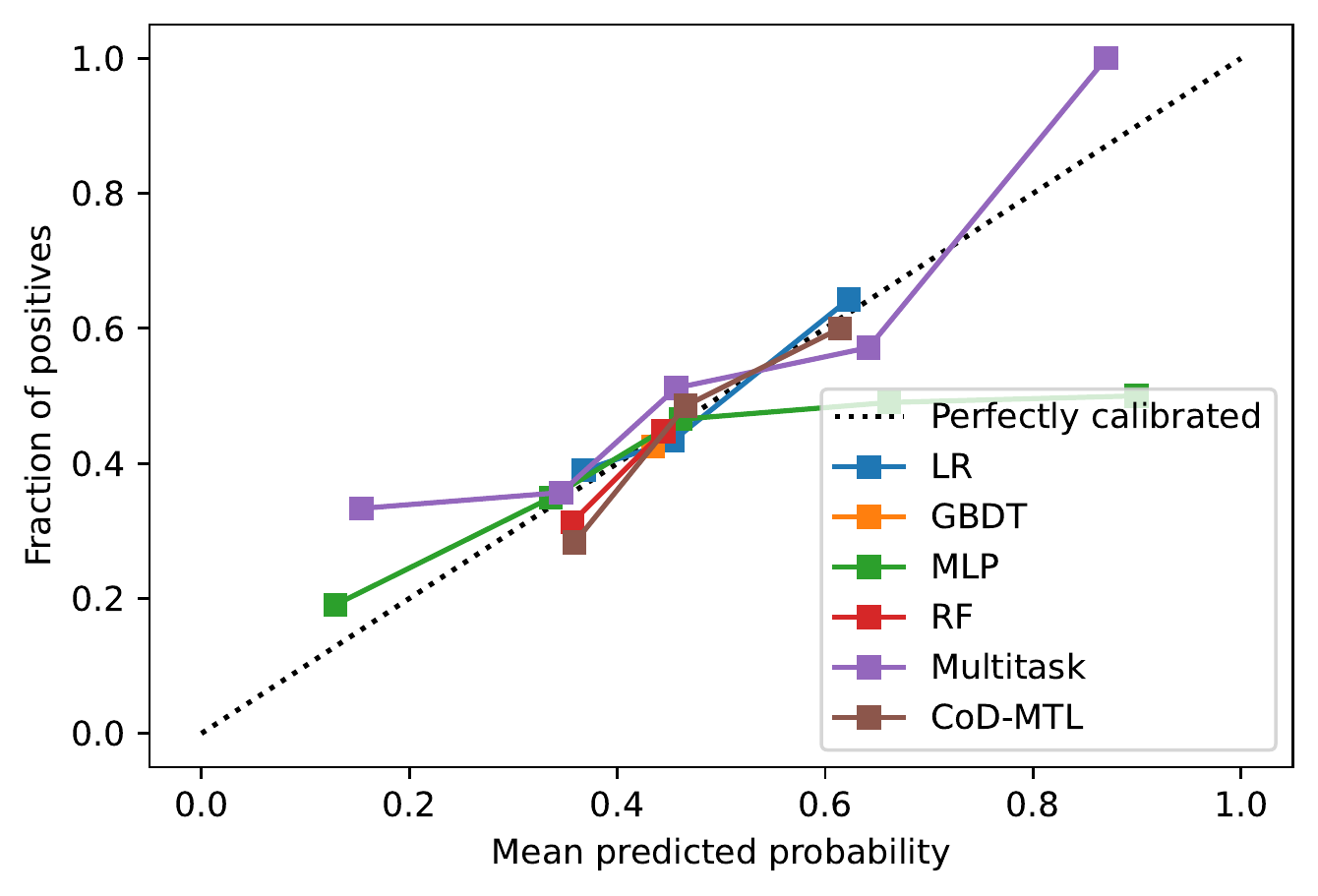}
    \includegraphics[scale=0.6]{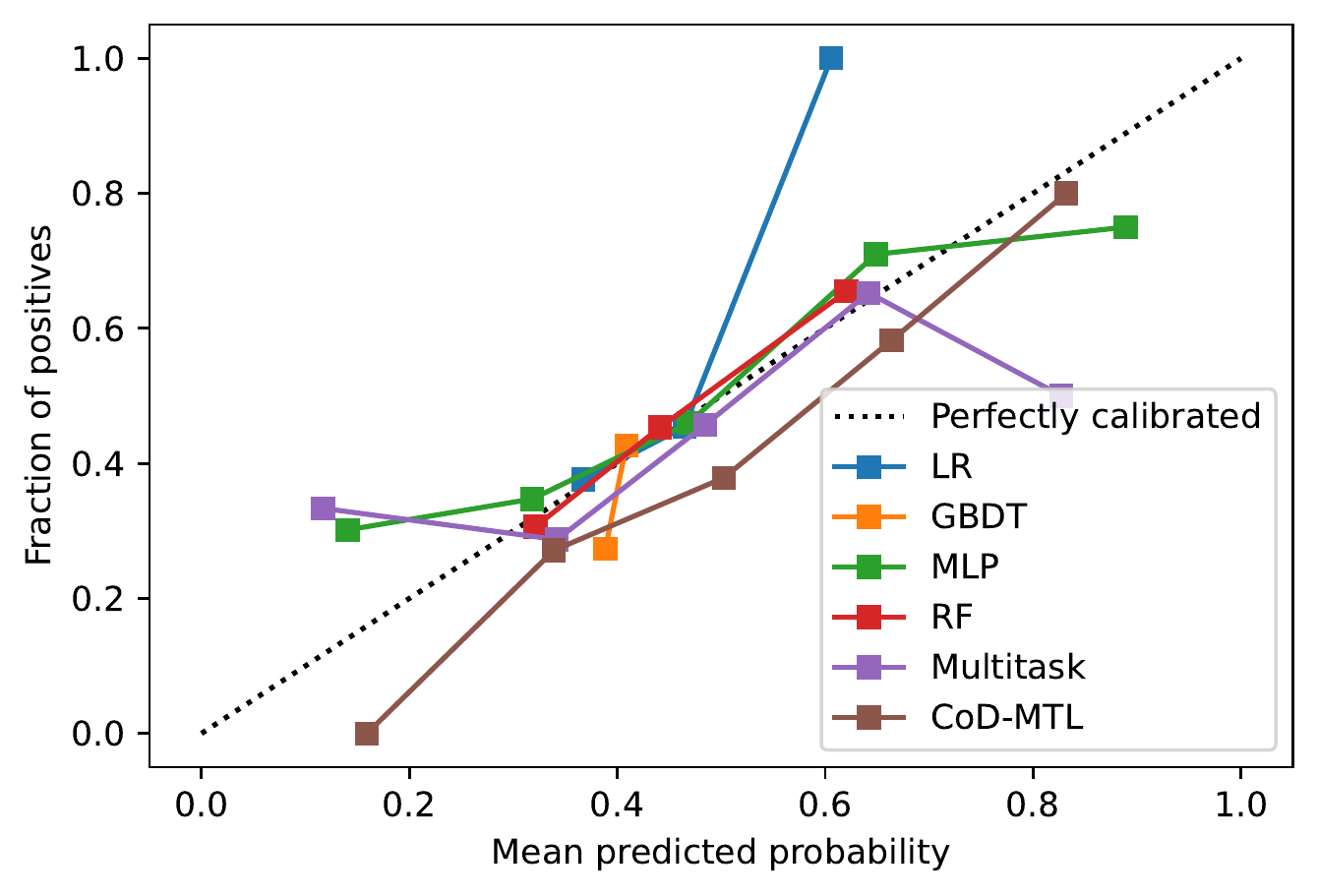}
 \caption{Calibration curves for rejection and infection CoDs (From left to right)}
 \label{fig:calib}
\end{figure}

In this section, we provide a comprehensive evaluation of CoD-MTL from the computational and clinical perspectives by answering the following research questions (\textbf{RQ}).

\begin{itemize}
    \item \textbf{RQ1:} Can the CoD-MTL accurately predict the rejection and infection as the CoDs? (Section~\ref{prediction})
    \item \textbf{RQ2:} To what extent can the Cod-MTL be considered trustworthy for predicting CoDs? (Section~\ref{uncertainty})
    \item \textbf{RQ3:} How could the CoD-MTL help the doctor make the clinical decision in liver transplant? (Section~\ref{casestudy})
\end{itemize}


\subsection{Experimental settings}
\textbf{Baseline methods.} We choose the baseline methods from three categories which are traditional ML, neural network, and multitask learning model respectively. For traditional ML, we select three commonly used methods as baselines, which are Logistic Regression (LR)~\cite{fan2008liblinear}, Gradient Boosting Decision Tree (GBDT), and Random Forest (RF)~\cite{breiman2001randomforest}. For the neural network, we use a multi-layer perceptron (MLP)~\cite{shickel2017deep} as the neural network baseline model. For the multitask learning model, we use the hard parameter sharing multitask learning frameworks as the baseline method.

\textbf{Evaluation metrics.} To ensure a fair comparison, we adopt the K-fold cross-validation strategy to evaluate the baseline and proposed methods. The AUROC and AUPRC metrics will be computed by averaging across multiple folds to assess the prediction accuracy. Additionally, we calculate the standard deviation (STD) of AUROC$/$AUPRC across different folds to evaluate the model uncertainty. To evaluate the clinical significance of CoD-MTL, we will engage a clinical expert to assist us in the case study.

\textbf{Implementation details.} We implemented the baseline machine learning methods using scikit-learn~\cite{scikit-learn} and PyTorch. The CoD-MTL framework was implemented using LightGBM~\cite{ke2017lightgbm} and PyTorch. We trained the CoD-MTL for $100$ epochs using AdamW as the optimizer with a learning rate of $0.001$. All the experiments were conducted on a server equipped with NVIDIA V100 GPUs and Intel Xeon CPUs. We set K to $4$ for cross-validation.

\begin{figure}[]
  \centering
    \includegraphics[scale=0.53]{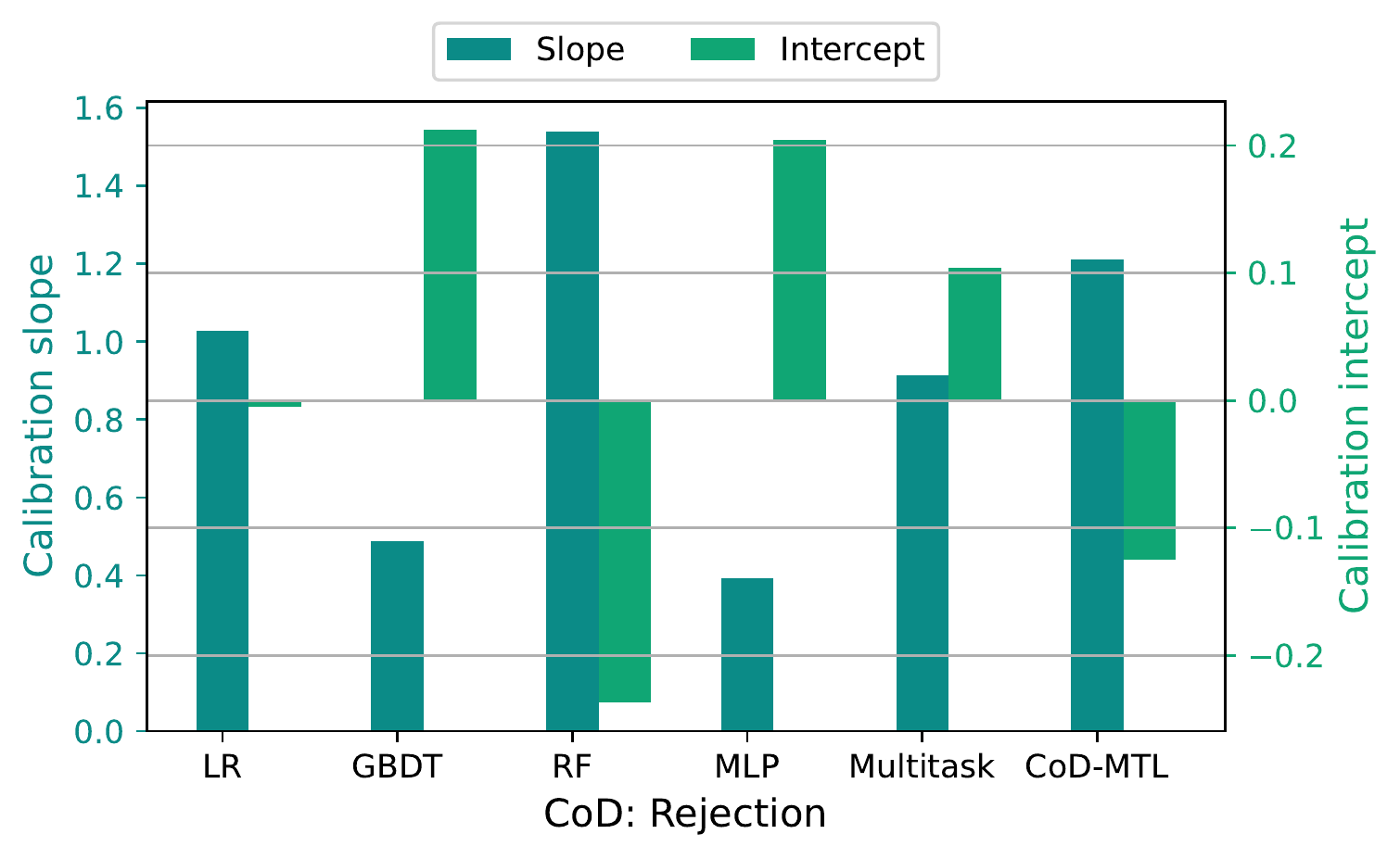}
    \includegraphics[scale=0.53]{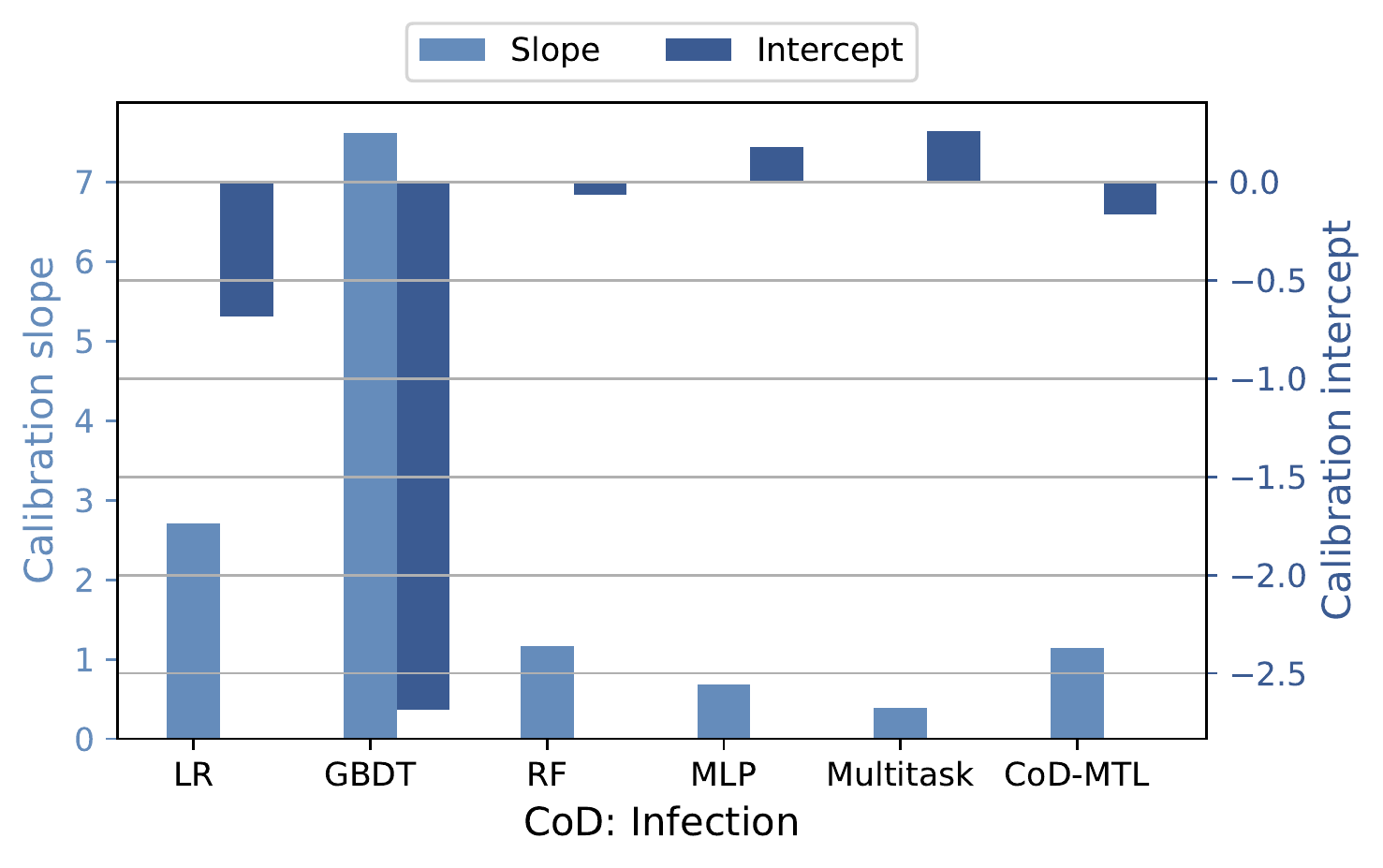}
 \caption{Calibration performance on rejection and infection prediction tasks.}
 \label{fig:slopeintercept}
\end{figure}

\subsection{Prediction performance on rejection and infection as CoDs}
\label{prediction}

We present the superior performance of CoD-MTL compared to the baseline machine learning methods as shown in Table~\ref{mainexp}. Several observations can be summarized as follows:

Firstly, the tree model outperforms MLP method on the CoD prediction task. For the rejection prediction, we observe that GBDT can achieve higher AUROC and AUPRC by $3.0\%$ and $0.81\%$ compared to MLP. For the infection prediction task, the AUROC and AUPRC of GBDT are higher than MLP by $3.2\%$ and $3.3\%$ respectively. This may be due to the ability of GBDT to identify important features from the EHR and eliminate irrelevant features that are less related.

Secondly, the multitask learning model improves performance compared to the single MLP. For the rejection prediction, the multitask learning baseline outperforms the single MLP by $4.2\%$ and $4.9\%$ on AUROC and AUPRC, respectively. For the infection prediction, the multitask learning baseline achieves $3.7\%$ and $6.6\%$ higher AUROC and AUPRC than the single MLP model. Our findings demonstrate that combining two highly related tasks in the multitask learning model can boost the performance of each single task. The shared model parameters can help the model learn common knowledge for both tasks and make more precise predictions for each CoD of the patients.

Thirdly, our results demonstrate that CoD-MTL outperforms the other baseline methods by a significant margin. Specifically, we observe a maximum improvement of $16.1\%$ and $15.6\%$ in terms of AUROC for rejection and infection prediction, respectively. Similarly, the maximum improvement in AUPRC is $15.6\%$ and $17.4\%$ for rejection and infection prediction, respectively. These results provide strong evidence of the effectiveness of CoD-MTL in leveraging the advantages of both tree models and multitask learning. By utilizing highly related features and common knowledge between the two tasks, CoD-MTL achieves superior performance on both CoD prediction tasks.

Moreover, we performed the sensitivity analysis using the ROC curve for a single fold of data. Figure~\ref{fig:roccurve} shows the ROC curves for rejection and infection prediction tasks. As seen in the figure, CoD-MTL exhibits a steeper slope than other baseline methods for both tasks. This indicates that CoD-MTL has a higher sensitivity, which is crucial for accurately identifying patients at high risk of rejection and infection. Early detection of rejection and infection is critical for preventing organ failure or loss and timely medical intervention. Therefore, the superior sensitivity of CoD-MTL makes it a promising approach for liver transplant outcome prediction.

\begin{figure*}[t]
  \centering
    \includegraphics[width=1.0\textwidth]{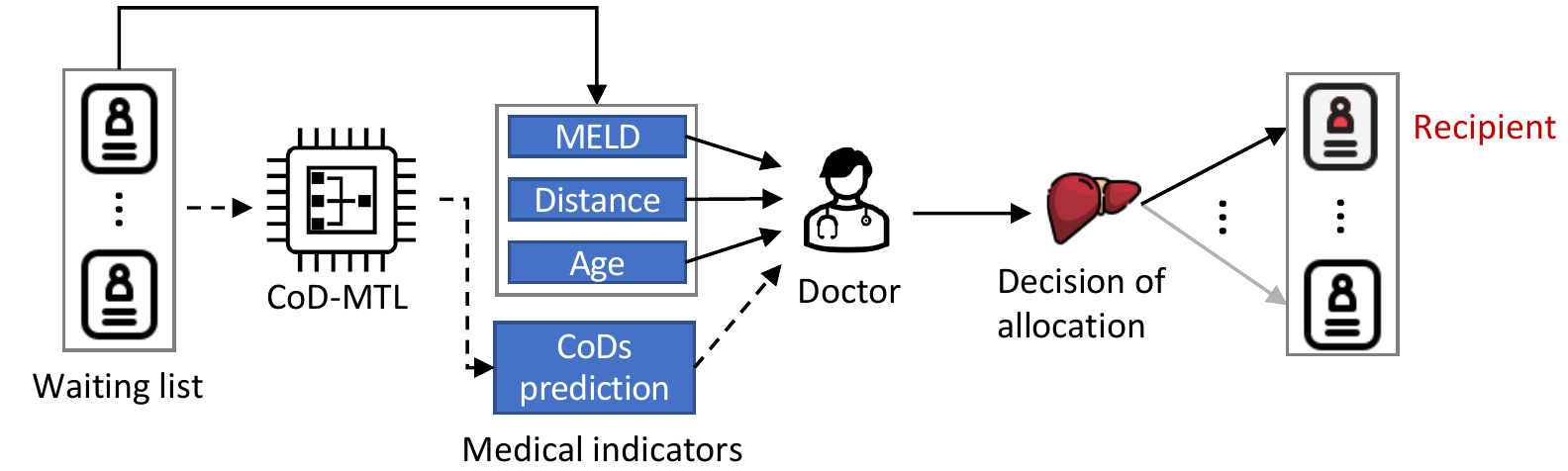}
 \caption{Illustration of how can CoD-MTL help the doctor make the clinical decisions in liver transplant.}
 \label{casestudyflow}
\end{figure*}

\subsection{Model uncertainty analysis}
\label{uncertainty}
Our proposed model shows moderately small uncertainty under cross-validation experiments, which is consistent with reliable machine learning models~\cite{baumann2014reliable}. We obtain the standard deviation (std) of the AUROC and AUPRC from cross-validation to evaluate the uncertainty. In the rejection prediction task, the CoD-MTL's std of AUROC is relatively small. Although LR and GBDT have smaller std than CoD-MTL, their AUROC is too low for precise prediction, which is $13.9\%$ lower than CoD-MTL. The CoD-MTL's std of AUPRC for rejection is fairly small. While LR reaches the smallest std of AUPRC, the AUPRC of LR is $15.6\%$ lower than CoD-MTL to ensure an accurate prediction. The situation is analogous when it comes to infection prediction. The std of AUROC of CoD-MTL is relatively small, and the only baseline method with a smaller std is MLP. However, MLP's AUROC is $9.1\%$ lower than CoD-MTL. MLP also has the lowest std of AUPRC, but its AUPRC is $14.5\%$ lower than CoD-MTL, indicating the incapability of infection prediction. The small std of our model shows its stable performance across different folds of data, which makes it promising to be generalized well to unseen data and reliably used in real-world practice.

Additionally, the proposed ML model can produce reliable predictions with well-calibrated probabilities, which is crucial for clinical applications~\cite{gawlikowski2021survey}. To further investigate the model uncertainty on rejection and infection prediction, we plot the calibration curves on one fold of data, as shown in Figure~\ref{fig:calib}. The calibration curve of CoD-MTL in both the left and right parts of Figure~\ref{fig:calib} is close to the diagonal line, indicating that the predicted probabilities correspond to the observed fractions well. To quantitatively measure the calibration performance of the models, we calculate the calibration slope and intercept of the calibration curve in Figure~\ref{fig:calib}, as shown in Figure~\ref{fig:slopeintercept}. The calibration slope of CoD-MTL is close to $1$ on both tasks, indicating that the predicted probabilities are well-calibrated with the true probabilities. Although LR's calibration slope is closer to $1$ on the rejection prediction task, it is not well-calibrated on the infection task. Similarly, the proposed model achieves a calibration intercept near $0$ on both tasks, indicating that the predicted probabilities are well-centered around the observed fractions. These observations suggest that the proposed CoD-MTL model can output reliable predictions with rather small uncertainty across different tasks, making it a promising tool for organ transplant outcome prediction.

\subsection{Case study}
\label{casestudy}
Our proposed model represents a significant improvement in clinical decision support for liver transplantation as shown in Figure~\ref{casestudyflow}. It takes into account a variety of factors that can affect patient outcomes, such as the likelihood of rejection or infection, to provide a more nuanced analysis of each patient's individual medical situation.

\begin{figure}[]
  \centering
    \includegraphics[scale=0.6]{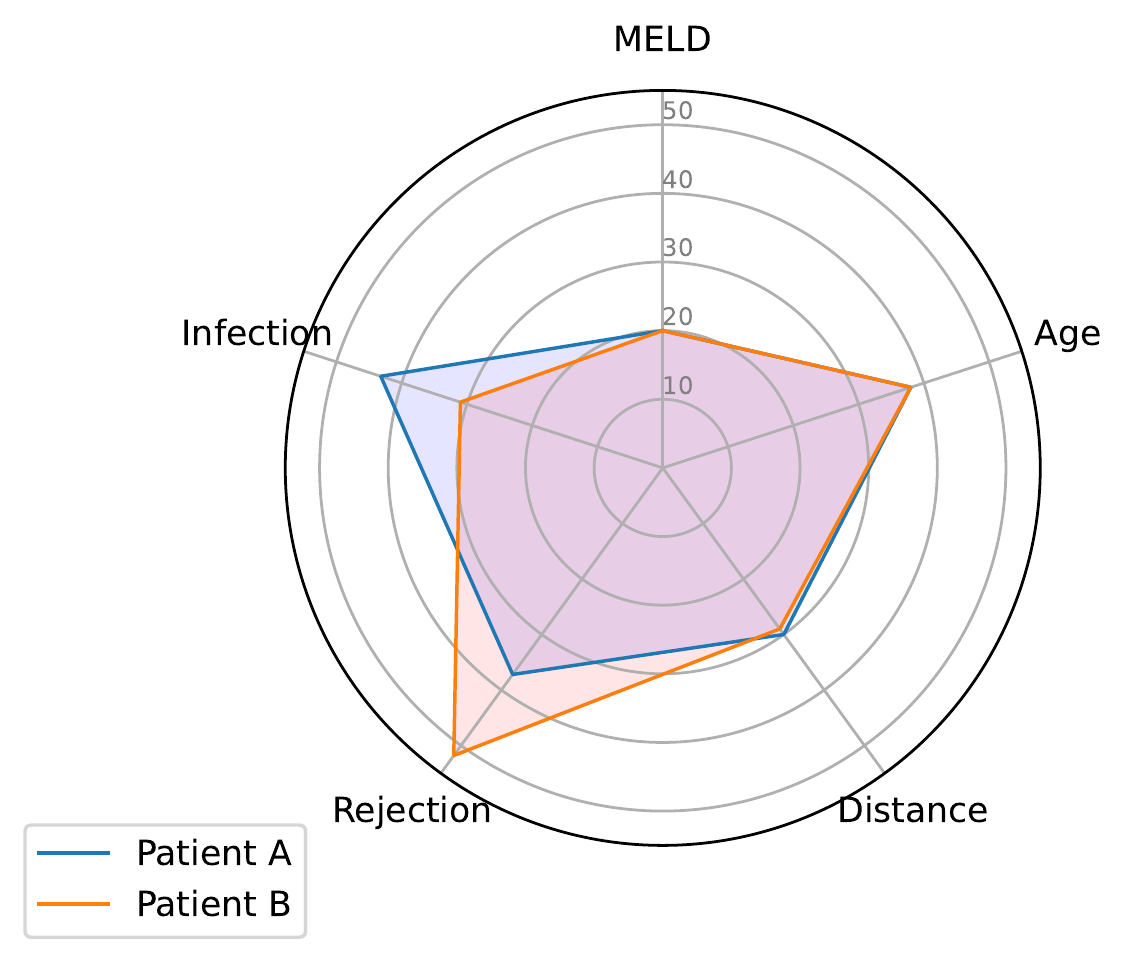}
    \includegraphics[scale=0.6]{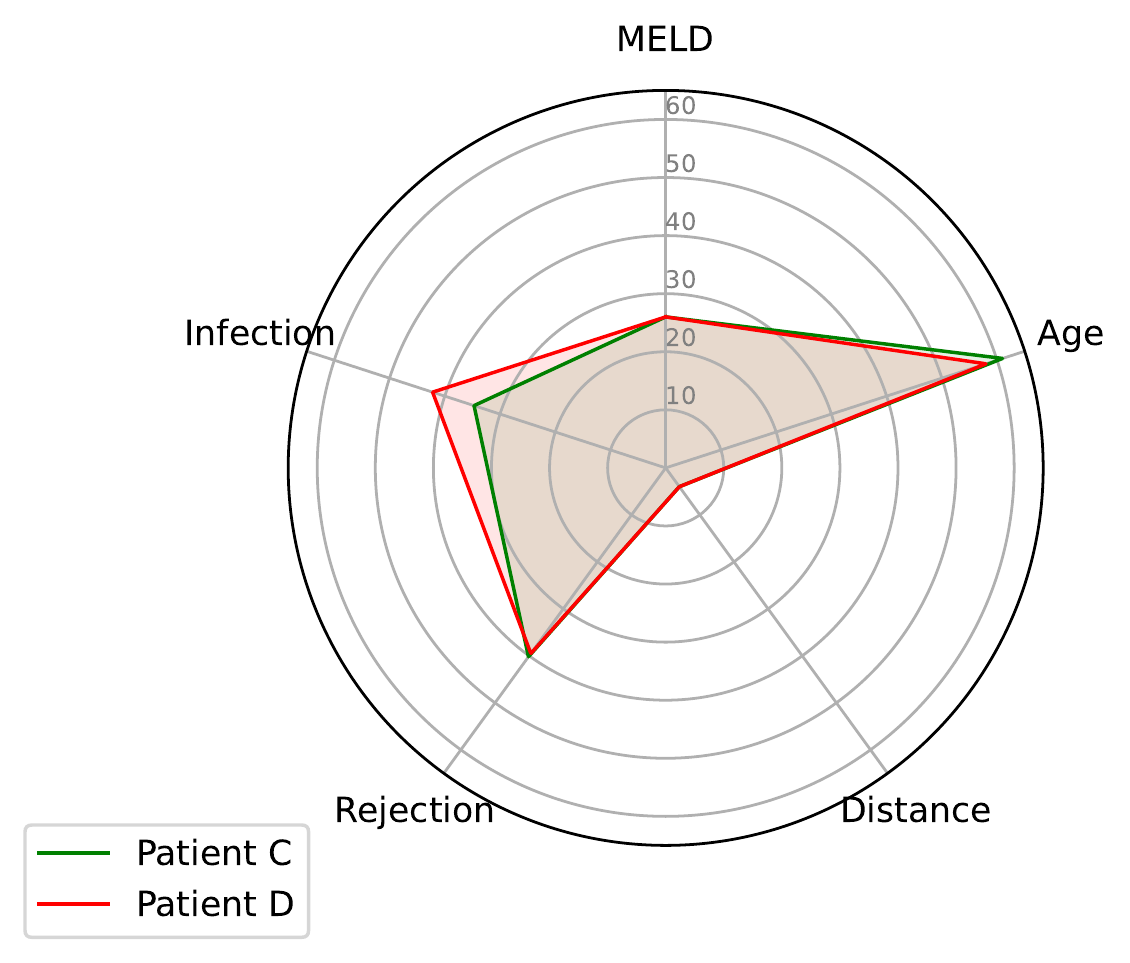}
 \caption{Two pairs of patients with similar features from the same transplant centers.}
 \label{radar}
\end{figure}

For instance, in cases where two patients from the same transplant center appear to be very similar, our model may reveal that they have different probabilities of dying from rejection or infection. We present two pairs of patients as shown in Figure~\ref{radar}. Patients A and B come from the same transplant center, and patients C and D come from another transplant center. We can observe that patients A and B have the same MELD score which is $20$, the same age which is $38$, and nearly the same distance from the donor which is $30$ and $29$ respectively. Our model predicts patient A with a higher probability of dying from infection and patient B with a higher probability of dying from rejection. The situation is similar for patients C and D, who share very similar characteristics related to allocation. Patient D is predicted with a higher probability of infection.

This kind of detailed analysis can be invaluable for clinicians who are looking to make more informed decisions about patient care~\cite{watt2010evolution}. With this level of information, doctors can develop more personalized treatment plans that are tailored to the specific needs of each patient. For example, they may choose to administer more aggressive immunosuppressant therapy to a patient who is at a higher risk of rejection, while opting for a more cautious approach for a patient who is at a lower risk. By providing doctors with more detailed and accurate information about patient outcomes, our model can help to improve the overall efficiency and effectiveness of liver transplantation. This, in turn, can lead to better outcomes for patients and more efficient use of healthcare resources.

\section{Discussion of limitation}

Additionally, we need to address some limitations and identify corresponding solutions for future improvement of our CoD-MTL framework. A critical constraint of our current model is its lack of interpretability, as the inability to explain predictions may impede its deployment in organ transplant scenarios. To address this gap, we will incorporate explainable AI techniques~\cite{tjoa2020survey} into our future work to provide a human-understandable interpretation of the predictions. Another limitation is the failure to consider fairness in our current framework. Equity is an ethical goal that clinical decision support systems should aim to achieve~\cite{rajkomar2018ensuring, chang2023towards}. Therefore, our future work will place a strong emphasis on integrating fairness constraints within our prediction framework to ensure it is suitable for multiple outcomes in organ transplants. The proposed framework has the potential to extend beyond our focus on organ transplants and can be applied to other medical fields that use multi-task learning. For instance, the prediction of complications and the length of stay in the ICU~\cite{harutyunyan2019multitask,zhang2023pheme} may benefit from our CoD-MTL framework, subject to future refinement.

\section{Conclusion}
\vspace{-0.2cm}
In this work, we propose a novel multi-task learning framework named CoD-MTL for the cause of death prediction in organ transplant. The key innovation lies in designing a tree-distillation strategy in multi-task learning, which serves as a bridge to combine the merits of tree-based models and multi-task deep neural networks for more accurate prediction of the transplant EHR data.
Empirical results on the liver transplant cohort show the output of CoD-MTL to be accurate and reliable for the precise liver transplant. The clinical case study further demonstrates our framework can be a promising clinical decision support tool for physicians in organ transplantation-related allocation and treatment procedures. We will attach more emphasis on the explainability and fairness of the framework as a future direction.

\section{Acknowledgements}
XJ is CPRIT Scholar in Cancer Research (RR180012), and he was supported in part by Christopher Sarofim Family Professorship, UT Stars award, UTHealth startup, the National Institute of Health (NIH) under award number R01AG066749, R01AG066749-03S1, R01LM013712, U24LM013755 and U01TR002062, and the National Science Foundation (NSF) \#2124789.

\makeatletter
\renewcommand{\@biblabel}[1]{\hfill #1.}
\makeatother

\bibliographystyle{vancouver}
\bibliography{amia}  

\end{document}